\documentclass[10pt,twocolumn,letterpaper]{article}

\usepackage{cvpr}              

\usepackage{graphicx}
\usepackage{amsmath}
\usepackage{amssymb}
\usepackage{booktabs}

\usepackage[utf8]{inputenc} 
\usepackage[T1]{fontenc}    
\usepackage{url}            
\usepackage{booktabs}       
\usepackage{amsfonts}       
\usepackage{nicefrac}       
\usepackage{microtype}      
\usepackage{xcolor}         
\usepackage{soul}

\usepackage{times}
\usepackage{epsfig}
\usepackage{graphicx}
\usepackage{amsmath}
\usepackage{amssymb}
\usepackage{multirow}
\usepackage{unicode}

\usepackage{subcaption,booktabs}

\usepackage{adjustbox}

\usepackage{amsfonts}       

\usepackage{times}
\usepackage{epsfig}
\usepackage{graphicx}
\usepackage{amsmath}
\usepackage{amssymb}
\usepackage{multirow}
\usepackage{color}
\usepackage{algorithmic}
\usepackage{algorithm}
\usepackage{booktabs}
\usepackage{url}

\usepackage{colortbl}
\usepackage{bm}
\usepackage{subcaption}
\usepackage{makecell}
\usepackage{pifont}
\usepackage{enumitem}
\usepackage{wrapfig}

\usepackage[title]{appendix}

\usepackage{url}

\usepackage{xcolor}

\usepackage[pagebackref,breaklinks,colorlinks]{hyperref}

\usepackage[capitalize]{cleveref}
\crefname{section}{Sec.}{Secs.}
\Crefname{section}{Section}{Sections}
\Crefname{table}{Table}{Tables}
\crefname{table}{Tab.}{Tabs.}

\def\cvprPaperID{10227} 
\def\confName{CVPR}
\def\confYear{2022}

\begin{document}

\title{Hyper-relationship Learning Network for Scene Graph Generation}

\author{Yibing Zhan \textsuperscript{\rm 1},
	Zhi Chen \textsuperscript{\rm 2},
	Jun Yu \textsuperscript{\rm 2},
	BaoSheng Yu \textsuperscript{\rm 3},
	Yong Luo \textsuperscript{\rm 4},
	Dacheng Tao \textsuperscript{\rm 1}\\
	\textsuperscript{\rm 1}JD Explore Academy,China,
	\textsuperscript{\rm 2}Hangzhou Dianzi University,
	\textsuperscript{\rm 3}The University of Sydney, Australia,\\
	\textsuperscript{\rm 4} National Engineering Research Center for Multimedia Software, Institute of \\Artificial Intelligence, School of Computer
	Science and Hubei Key Laboratory of\\ Multimedia and Network Communication Engineering, Wuhan University, China,\\
	zhanyibing@jd.com,\{zhixiao996,yujun\}@hdu.edu.cn,luoyong@whu.edu.cn,\\  baosheng.yu@sydney.edu.au, dacheng.tao@gmail.com.
}
\maketitle

\begin{abstract}
%
  Generating informative scene graphs from images requires integrating and reasoning from various graph components, i.e., objects and relationships. However, current scene graph generation (SGG) methods, including the unbiased SGG methods, still struggle to predict informative relationships due to the lack of 1) high-level inference such as transitive inference between relationships and 2) efficient mechanisms that can incorporate all interactions of graph components. To address the issues mentioned above, we devise a hyper-relationship learning network, termed HLN, for SGG. Specifically, the proposed HLN stems from hypergraphs and two graph attention networks (GATs) are designed to infer relationships: 1) the object-relationship GAT or OR-GAT to explore interactions between objects and relationships, and 2) the hyper-relationship GAT or HR-GAT to integrate transitive inference of hyper-relationships, i.e., the sequential relationships between three objects for transitive reasoning. As a result, HLN significantly improves the performance of scene graph generation by integrating and reasoning from object interactions, relationship interactions, and transitive inference of hyper-relationships. We evaluate HLN on the most popular SGG dataset, i.e., the Visual Genome dataset, and the experimental results demonstrate its great superiority over recent state-of-the-art methods. For example, the proposed HLN improves the recall per relationship from 11.3\% to 13.1\%, and maintains the recall per image from 19.8\% to 34.9\%. We will release the source code and pretrained models on GitHub.


\end{abstract}

\vspace{-0.4cm}
\begin{figure}[t]
{\includegraphics[width=1.0\linewidth]{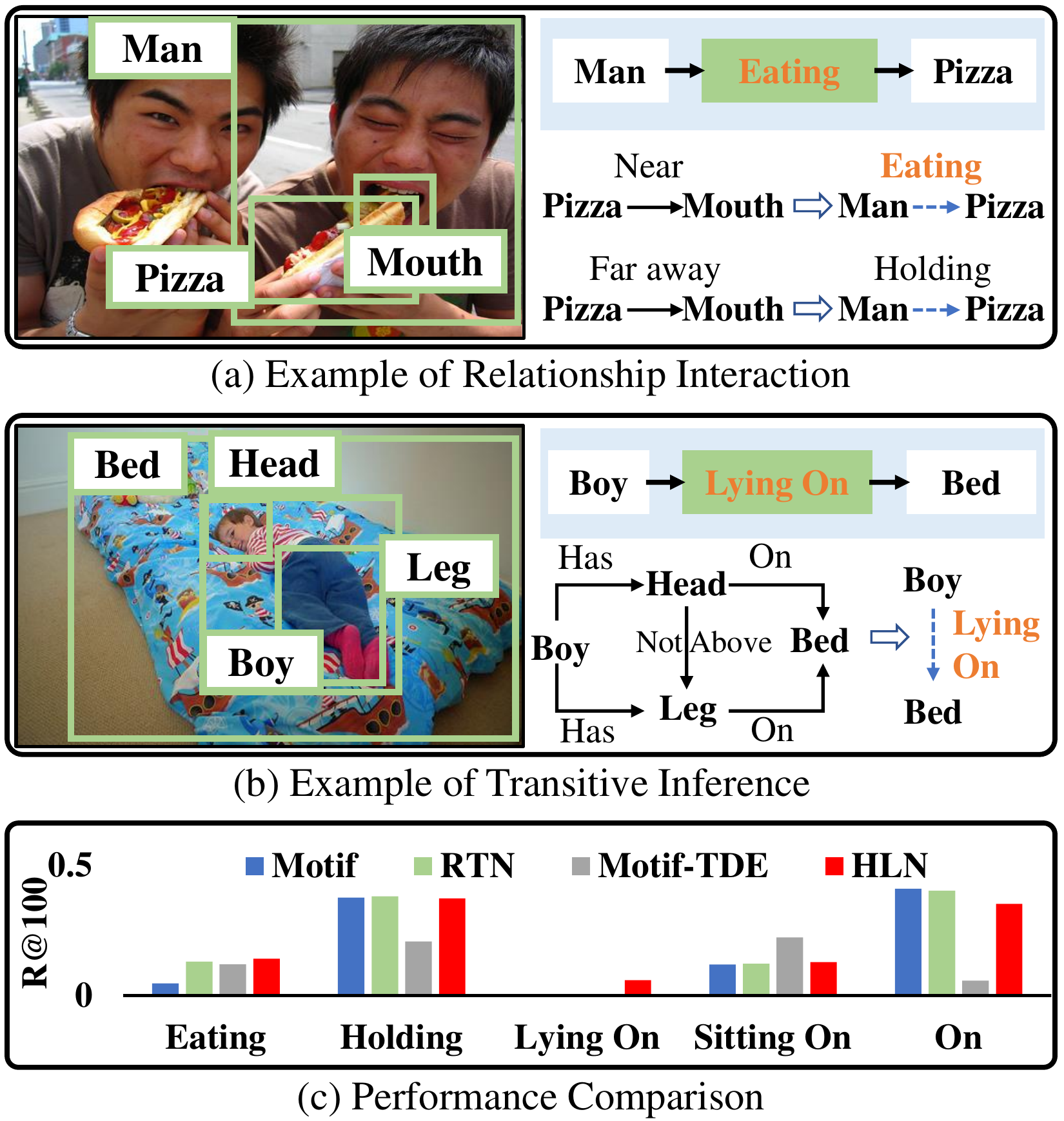}}
\caption{Examples of exploring interaction and transitive inference using the representative SGG methods: Motif \cite{zellers2018neural}, RTN \cite{koner2020relation}, Motif-TDE \cite{tang2020unbiased}, and our proposed HLN. Specifically, Motif, RTN, and HLN use object interactions, object and relationship interactions, and all connections, respectively. Motif-TDE is a typical unbiased SGG. (a) Motif provides ``holding'' and other methods provide ``eating''. (b) HLN provides ``lying on'', Motif-TDE provides ``sitting on'', and other methods provide ``on''. (c) the Recall@100 performance using the relationships in (a) and (b) on the Visual Genome dataset \cite{krishna2017visual}. The above examples demonstrate the effectiveness of interactions and transitive inference for SGG.}
\label{fig1}
\end{figure}
\section{Introduction}\label{sec:intro}

Scene graph generation (SGG) aims to detect objects and predict object relationships. These detected objects and relationships then constitute the scene graphs of images. The generated scene graphs provide not only visual content for image understanding but also knowledge representation to benefit high-level visual applications, such as image captioning \cite{yao2018exploring} and visual question answering \cite{teney2017graph}. The key of SGG is to model and explore the connections between the objects and the object relationships~\cite{xu2017scene}. However, most current SGG methods only take advantage of object interactions \cite{zellers2018neural,woo2018linknet,qi2019attentive,chen2019counterfactual,gu2019scene,chen2019knowledge,lin2020gps}, leaving the intrinsic connections of relationships poorly investigated. For instance, these methods cannot understand the relationship interaction in Fig. \ref{fig1} (a): comparing with ``Holding'',  ``Eating'' is a better description of the relationship between the man and the pizza, since the pizza is close to the mouth.   

Several recent works have explored relationship interactions for SGG \cite{xu2017scene,yang2018graph,wang2019exploring,koner2020relation,ren2020scene,zhang2020dual}. However, high-level connections of relationships, \ie, transitive inference \cite{gillan1981reasoning,acuna2002frontal,vasconcelos2008transitive,lazareva2012transitive}, are usually ignored in these methods, thus lacking the ability to deal with more informative relationships.
Transitive inference refers to the inference of the relationship between two objects by consolidating the relationships between the two objects and another mediating object. Exploiting transitive inference makes better structures and integration for the surrounding relationships. As shown in Fig. \ref{fig1} (b), the boy's head is on the bed is more meaningful than the head is on the bed when inferring the relationship between the boy and the bed. The generation of ``Boy-Lying On-Bed'' requires to integrate multiple relationships, including 1) the boy's head and leg on the bed and 2) the boy's head is not above the boy's leg. 

Recent studies on unbiased SGG blamed the above relationship prediction problems on the bias of the training set \cite{tang2020unbiased,yan2020pcpl,wang2020memory,wang2020tackling}. These unbiased studies proposed debiasing strategies, which still ignored interaction and inference and did not improve the relationship detection ability. In Fig. \ref{fig1} (c), we compare the performance of four types of SGG methods: Motif \cite{zellers2018neural}, Motif-TDE \cite{tang2020unbiased}, RTN \cite{koner2020relation}, and our proposed HLN. Motif, RTN, and HLN are representative methods that use object interactions, object and relationship interactions, and all types of connections, respectively. Motif-TDE is a representative unbiased SGG method based on Motif. In Fig. \ref{fig1} (c), Motif-TDE yields higher detection of less-frequently seen relationships ``sitting on/eating'', but by severely sacrificing the detection of more-frequently seen relationships ``on/holding''. 

In light of the above analysis, we develop a hyper-relationship learning network (HLN) to explore and exploit connections of objects and relationships for SGG. The main contributions of this paper lie in three aspects:

First, we exploit hypergraphs and propose hyper relationships to naturally and seamlessly integrate interaction and transitive inference. The hyper relationships are defined as the subsets of relationships between three objects to model transitive reasoning \cite{lazareva2012transitive}. As far as we know, HLN is the first work exploring transitive inference for SGG.

Second, we develop an object-relationship graph attention network (OR-GAT) to incorporate interaction between objects and relationships. Specifically, OR-GAT first passes information from relationships to objects and then collects information from objects to relationships. Such a manner implicitly considers most types of graph component interactions with lower complexity.

Third, we design a hyper-relationship GAT (HR-GAT) to model the transitive inference for SGG.  Specifically, for each relationship, HR-GAT first collects the transitive inference from the corresponding hyper relationships and then integrates the collected transitive inference to the corresponding relationship in an attentional manner. By using HR-GAT, HLN has abilities to understand the high-level connections between graph components.

The proposed HLN consists of three modules: an object proposal network (OPN), an object classifier, and a relationship predictor. Specifically, we use Faster R-CNN \cite{ren2015faster} as the OPN for fair comparisons with \cite{tang2020unbiased,wang2020tackling}. The object classifier comprises several Transformer layers \cite{vaswani2017attention} to exploit object interactions. The relationship predictor contains one OR-GAT followed by one HR-GAT to combines object interactions, relationship interactions, and transitive inference. Comprehensive experiments were conducted on the Visual Genome \cite{krishna2017visual} dataset. The performance compared with the state-of-the-art demonstrates that HLN is capable of detecting various types of relationships. In particular, HLN can detect informative relationships, such as ``Playing/Painted On'', which are hardly recognized by previous SGG methods, if graph constraints \cite{zellers2018neural} are not considered. The ablation experiments validate the benefits of interaction and transitive inference and the superiority of the designed OR-GAT and HR-GAT for SGG.

\section{Related Works}
 \begin{figure*}[t]
 \centering
{\includegraphics[width=1\linewidth]{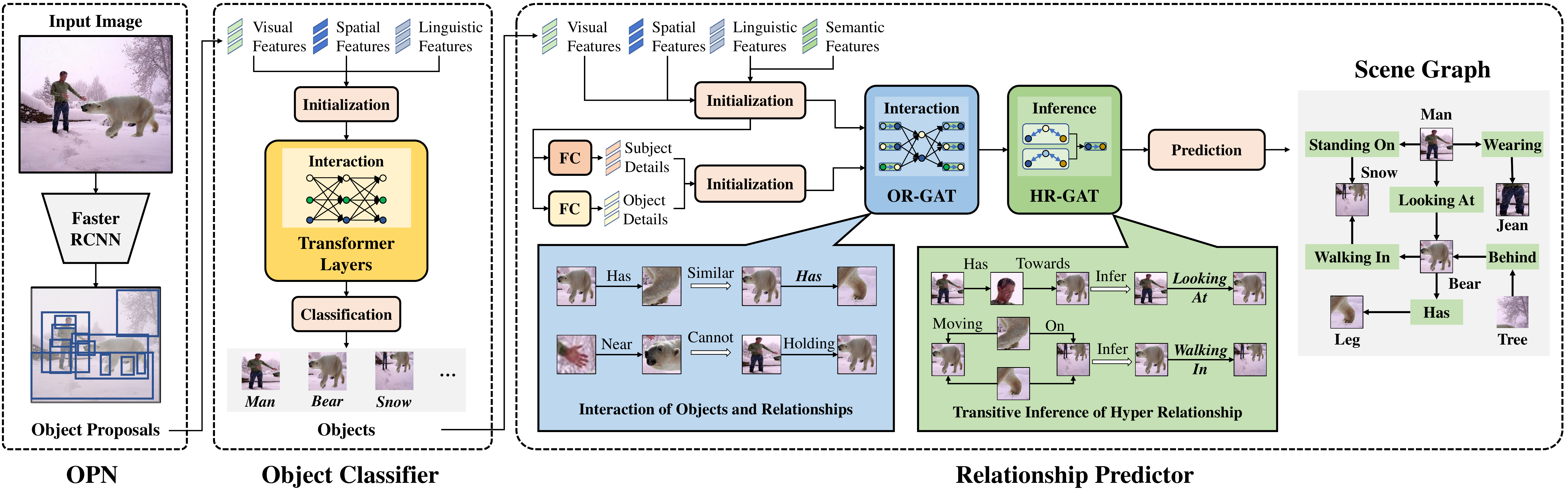}}
\caption{The framework of HLN. HLN consists of three modules: an object proposal network (OPN), an object classifier, and a relationship predictor. Specifically, HLN first obtains object proposals based on the OPN. Then, objects are detected by using the object classifier to exploit the object interaction. Afterward, every two objects constitute a relationship proposal. The relationship predictor finally predicts the relationships based on object interaction, relationship interaction, and transitive inference.}
\label{fig2}
\end{figure*}
\textbf{Scene Graph Generation.} Earlier works on SGG refers to visual relationship detection (VRD), which detects the relationship between two objects~\cite{lu2016visual,zhang2017visual,zhan2019exploring,zhang2019graphical,zhan2020multi}. Later, SGG methods considered the surrounding contexts in whole images \cite{xu2017scene}. Recent methods have exploited object interactions using either bidirectional LSTMs \cite{huang2015bidirectional,zellers2018neural,qi2019attentive} or self-attention networks \cite{vaswani2017attention}. Furthermore, \cite{li2018factorizable} and \cite{gu2019scene} clustered neighboring objects as subgraphs and then passed information between subgraphs and objects. Nevertheless, the above methods usually ignored relationship connections. 

Several recent works have explored the relationship interaction. For example, \cite{xu2017scene}, \cite{yang2018graph}, and \cite{wang2019exploring} updated objects using their constituted relationships and updated relationships using two corresponding objects. However, the above methods probably missed valuable interactions between objects and relationships that are not constituted by the objects \cite{ren2020scene}. \cite{zareian2020weakly} updated relationships using all objects. However, in \cite{zareian2020weakly}, relationships were randomly initialized during training, seriously affecting the performance. \cite{koner2020relation}, \cite{ren2020scene}, and \cite{zhang2020dual} calculated relationship interaction between all relationships. Nevertheless, $N$ objects result in $N^2$ relationships. The above methods had a high computational complexity. Besides, no SGG methods have explore the transitive inference. Recently, unbiased SGG has been a popular topic. \cite{tang2020unbiased} used causal inference to re-adjust the relationship representations. \cite{yu2020cogtree} introduced cognition tree loss to re-optimize the prediction. However, existing unbiased SGG strategies did not explore relationship connections and performed weakly in detecting informative relationships.

\textbf{Visual Reasoning.} In SGG, visual reasoning plays a significant role in understanding the relationships between graph components. For example, \cite{koner2020relation} proposed a relation Transformer \cite{vaswani2017attention}, which considers relationships during implicit visual reasoning. \cite{gu2019scene} and \cite{tang2020unbiased} explicitly used common-sense knowledge and causal inference, respectively. HLN incorporates both objects and relationships during the implicit reasoning process and explicitly uses transitive inference \cite{lazareva2012transitive} to improve visual reasoning ability.

\textbf{Graph Attention Networks.} 
GATs are attention models that operate in the graph domain \cite{velivckovic2017graph,zhou2018graph}. We also regard the Transformer \cite{vaswani2017attention} as an attention graph model. In SGG, the connections of graph components are generally unavailable. GAT is an automatic manner to highlight important information, meanwhile reducing the side effects of possible noise \cite{yang2018graph,woo2018linknet,lin2020gps}. In this paper, we design two novel GATs to explore interaction and transitive inference for SGG.

\vspace{-0.1cm}
\section{Hyper-relationship Learning Network}
Fig. \ref{fig2} presents the framework of HLN, which consists of three modules: an OPN, an object classifier, and a relationship predictor. Specifically, HLN first obtains object proposals using the OPN. Then, the object classifier predicts each object's label using Transformer layers based on object interactions. Afterward, in the relationship predictor, every two detected objects constitute a relationship proposal. The final relationships are predicted using OR-GAT to exploit interactions between objects and relationships and HR-GAT to explore transitive inference. In the remaining subsections, we first give the problem formulation for SGG based on the hypergraph. Then, we sequentially explain the OPN, the object classifier, and the relationship predictor.

\subsection{Problem Formulation}
Previous methods generally defined scene graphs as graphs containing object proposals, objects, relationship proposals, and relationships \cite{yang2018graph,zellers2018neural,tang2020unbiased}. However, the above definition can only describe the interaction between graph components. In HLN, we exploit hypergraphs and propose hyper relationships to model transitive reasoning and further combine the inference and interaction for SGG.

Specifically, suppose that a hypergraph $\mathcal{H}$ for SGG is defined as $\mathcal{H}$=$\{\mathcal{B},\mathcal{O},\mathcal{V},\mathcal{E},\mathcal{R}\}$. Here, $\mathcal{B}$=$\{b_i\}$ denote object proposal set, $\mathcal{O}$=$\{o_i\}$ denote object set, $\mathcal{V}$=$\{v_{ij}\}$ denote relationship proposal set, $\mathcal{E}$=$\{E_{ijk}\}$ denote hyper relationship set, and $\mathcal{R}$=$\{r_{ij}\}$ denote relationship set, hyper relationships, and relationships, respectively. $i,j$$\in$$[N]$. $N$ is the total number of objects. We define $[N]$=$\{1,2,...,N\}$. Each $v_{ij}$ and $r_{ij}$ are constituted by two objects: $o_i$ and $o_j$. Based on transitive reasoning \cite{lazareva2012transitive}, the hyper relationship $E_{ijk}$ is defined as a subset of relationships $E_{ijk}$=$\{v_{ij}, v_{ji}, v_{ik},v_{ki},v_{jk},v_{kj}\}$ between three objects $o_i$, $o_j$, and $o_k$. The transitive inference of $E_{ijk}$ to $v_{ij}$ is calculated by integrating the subsets of relationships $\{v_{ik},v_{ki},v_{jk},v_{kj}\}\in E_{ijk}$. The problem formulation for generating a scene graph $\mathcal{G}$=$\{\mathcal{O},\mathcal{R}\}$$\in$$\mathcal{H}$ of image $I$ is calculated based on four factors:
\begin{equation}
\small
 P(\mathcal{G}|I)=  P(\mathcal{B}|I)P(\mathcal{O}|\mathcal{B},I) P(\mathcal{V}|\mathcal{O},\mathcal{B},I)P(\mathcal{R}|\mathcal{E},\mathcal{V},\mathcal{O},\mathcal{B},I). 
\end{equation}

The object proposal component $P(\mathcal{B}|I)$ and object component $P(\mathcal{O}|\mathcal{B}, I)$ is completed by the OPN and the object classifier, respectively. Both relationship proposal component $P(\mathcal{V}|\mathcal{O},\mathcal{B},I)$ and relationship component $P(\mathcal{R}|\mathcal{E},\mathcal{V},\mathcal{O},\mathcal{B},I)$ are modeled in the relationship predictor. $P(\mathcal{V}|\mathcal{O},\mathcal{B},I)$ is used to generate the relationship proposals based on the objects that constitute the corresponding relationship proposals. $P(\mathcal{R}|\mathcal{E},\mathcal{V},\mathcal{O},\mathcal{B},I)$ is used to detect relationships by combining object interaction, relationship interaction, and transitive inference of hyper relationships. 

\subsection{Object Proposal Network}
We use Faster R-CNN as the OPN following \cite{zellers2018neural,chen2019counterfactual,ren2020scene}. For one image $I$, the OPN generates a set of object proposals $\mathcal{B}$=$\{b_i\}$, $i$$\in$$[N]$. For each proposal $b_i$, OPN provides a spatial feature $\vec{p}_i \in \mathbb{R}^{9}$, a visual feature $\vec{v}_i\in \mathbb{R}^{d_v}$, and an object label probability $\vec{c}_i \in \mathbb{R}^{c_o + 1} $. The spatial feature $\vec{p}_i$ includes relative bounding box coordinates $(\frac{x_{i1}}{w}, \frac{y_{i1}}{h}, \frac{x_{i2}}{w}, \frac{y_{i2}}{h})$, relative center ($\frac{x_{i1}+x_{i2}}{2w}$,$\frac{y_{i1}+y_{i2}}{2h}$), and relative sizes ($\frac{x_{i2}-x_{i1}}{w}$,$\frac{y_{i2}-y_{i1}}{h}$, $\frac{(x_{i2}-x_{i1})(y_{i2}-y_{i1})}{wh}$). Here, ($x_{i1}, y_{i1}, x_{i2}, y_{i2}$) are the bounding box coordinates of $b_i$. $w$ and $h$ are the image width and height. $d_v$ is the dimension of visual feature. $c_o+1$ is the number of object categories, with the addition of an background category.

\subsection{Object Classifier}
We use the Transformer layers as the object classifier because it is capable of modeling the interaction between input objects. We do not consider relationship connections during object classification for computational efficiency \cite{yu2020cogtree}. 
Suppose an object proposal set $\mathcal{B}$=$\{b_i\}$, $i$$\in$$[N]$ is given. The feature $\vec{x}_i$ of $b_i$ is initialized by fusing the corresponding spatial features, visual features, and linguistic features: $\vec{x}_i=\sigma(\text{FC}_o(\vec{v}_i\|\vec{p}_i\|\text{Emb}_o(\vec{c}_i)))$,
where $\sigma$ is activation function. $\text{FC}(\vec{x})$=$W\vec{x}+b$ represents a linear transformation, $W$ and $b$ are the weight matrix and bias. ``$\|$'' indicates concatenation. $\text{Emb}_o(\cdot)$ indicates the linguistic embedding based on the GloVe model \cite{pennington2014glove}, following \cite{tang2019learning,tang2020unbiased}.

Let $X$$\in$$\mathbb{R}^{N\times d_o}$ denote the feature set of $\mathcal{B}$. $d_o$ is the feature dimension of $\vec{x_i}$. The process of using one Transformer Layer to update the object features is calculated as: 
\begin{equation}
\small
X'= \text{FFN}_o(\text{AT}_o(X,X,X)),
\end{equation}
where $\text{AT}_o(\cdot)$ indicates am attention network for exploring object interactions. The attention process is defined as:
\begin{equation}
\small
\text{AT}_o(X_1,X_2,X_3)=\text{Softmax}(
\frac{\text{Q}_{o}(X_1)\text{K}_{o}(X_2)^T}{\sqrt{d_{k_o}}})\text{V}_{o}(X_3),
\end{equation}
where $\text{Q}_o(\cdot)$, $\text{K}_o(\cdot)$, and $\text{V}_o(\cdot)$ are parallel linear transformations, representing the query, key, and value of the attention process, respectively. $d_{k_o}$ is the dimension of $\text{Q}_o(\cdot)$ and $\text{K}_o(\cdot)$. $\text{FFN}_o(\cdot)$ is a feed-forward network (FFN) \cite{vaswani2017attention}. 

The final objects are classified by using a softmax cross-entropy loss following \cite{tang2020unbiased}.

\subsection{Relationship Predictor}
Previous SGG methods lack the ability to comprehensively explore all connections (interaction and inference) between graph components for SGG. Consequently, previous SGG methods' detection of informative relationships is still limited, no matter the relationship is less or more frequently seen in the training dataset. We introduce the proposed HLN, including OR-GAT and HR-GAT, and explain how HLN combines 1) interactions of objects and relationships and 2) transitive inference to infer relationships.

We first introduce the object and relationship initialization. Given $N$ detected objects $\mathcal{O}$=$\{o_i\}$ along with the corresponding $N$ object proposals $\mathcal{B}$=$\{b_i\}$, $i$$\in$$[N]$, the object feature $\vec{y}_i$ of $o_i$ is then initialized as:
\begin{equation}
\vec{y_i}=\sigma(\text{FC}_{ro}(\vec{v}_i\|\vec{p}_i\|\vec{x}_i'\|\text{Emb}_r(\vec{l}_i))),
\end{equation}
where $\vec{v}_i$ and $\vec{p}_i$ are the visual and spatial feature of $o_i$, respectively. $\vec{x}_i'$ indicates the semantic feature obtained from the last layer of the Transformer layers of the object classifier. $\text{Emb}_r(\vec{l}_i)$ is the linguistic embedding \cite{pennington2014glove} based on the predicted one-hot label $\vec{l}_i$ of the object classifier. The feature $\vec{z}_{ij}$ of relationship proposal $v_{ij}$ between objects $o_i$ and $o_j$ is calculated as:
\begin{equation}
\vec{z}_{ij}=\sigma(\text{FC}_{v3}(\text{FC}_{v1}(\vec{y}_i)\|\text{FC}_{v2}(\vec{y}_j))).
\end{equation}

\textbf{OR-GAT.} Then, we introduce the object and relationship interaction. Typically, there are four types of interactions between graph components: one interaction between objects, one interaction between relationships, and two interactions between objects and relationships concerning the direction. The intuitive way is to explore the four types of interactions by four different modules. Nevertheless, there are two main problems: 1) $N$ objects result in $N^2$ relationships, and the direct calculation of interaction between relationships is thus computational costly; and 2) four different modules require a lot of parameters, and how to deeply incorporate the four types of interactions remains challenging. To address the above-mentioned issues, we design OR-GAT, consisting of two sequential attention networks, to incorporate four different types of interactions. Specifically, OR-GAT consists of two alternate GAT to pass information between objects and relationships. In such a manner, the interactions between objects and relationships are alternately modeled in OR-GAT, and object interaction and relationship interaction can thus be implicitly modeled by OR-GAT. 

We formulate the OR-GAT as follows: let $Y$$\in$$\mathbb{R}^{N\times d_o}$ and $Z$$\in$$\mathbb{R}^{N^2\times d_r}$ denote the feature sets of objects and relationship proposals, respectively. $d_r$ is the feature dimension of one relationship proposal. OR-GAT first updates object features by passing messages from relationships to objects:
\begin{equation}
Y'=\text{FFN}_{r1}(\text{MA}_{r1}(Y,Z,M_{r1})),
\end{equation}
where $\text{MA}_{r1}(\cdot)$ is a masked attention network, \ie,
\begin{equation}
\small
\text{MA}_{r1}(Y,Z,M_{r1})=\text{Softmax}(
\frac{\text{Q}_{r1}(Y)\text{K}_{r1}(Z)^TM_{r1}}{\sqrt{d_{k_{r1}}}})\text{V}_{r1}(Z),
\end{equation}
where $M_{r1}$ is a $N$$\times$$N^2$ mask to limit message passing only from relationships to their corresponding objects~\cite{yang2018graph}. This mask slightly improves the performance. Then, OR-GAT updates the relationship features by using the updated objects' messages as follows:
\begin{equation}
Z'=\text{FFN}_{r2}(\text{AT}_{r2}(Z,Y',Y')).
\end{equation}

\textbf{HR-GAT.} Modeling transitive inference is a complex problem, considering that: we are not aware of 1) which compositions of relationships in hyper relationships constitute a transitive inference, and 2) which transitive inference is useful for the targeted relationship. For simplicity, we use a fixed sequence of hyper relationships and propose HR-GAT to model the process. Given the relationship $r_{ij}$, HR-GAT first obtains the corresponding hyper-relationships $E_{ijk}$, $k$$\in$$[N]$. For transitive reasoning \cite{lazareva2012transitive}, the transitive inference from $E_{ijk}$ to $r_{ij}$ is calculated using the relationships between two objects $o_i$ and $o_j$ and the mediating object $o_k$. The transitive inference $\vec{h}_{ijk}$ can be defined as:
\begin{equation}
\vec{h}_{ijk}=\sigma(\text{FC}_h(\vec{z}_{ik}\| \vec{z}_{ki} \|\vec{z}_{jk}\| \vec{z}_{kj})).
\end{equation}

Note that for different relationships, the transitive inferences might be different. Let $H_{ij}$$\in$$\mathbb{R}^{N\times d_h}$ denote the transitive inference set of relationship $r_{ij}$. $d_h$ is the feature dimension. The transitive inference of $H_{ij}$ is integrated and passed to $r_{ij}$ using an attention method:
\begin{equation}
\vec{z}_{ij}' = \text{FFN}_h(\text{AT}_h(\vec{z}_{ij},H_{ij},H_{ij})).
\end{equation}

For robust performance, the multi-head strategy \cite{vaswani2017attention} is adopted in all attention networks. The residual connection with the layer normalization \cite{vaswani2017attention}, which is defined as $X$=$X+\text{LN}(\text{Fun}(X))$, is added to each attention network and each FFN. Here, $X$ is the input feature set, $\text{LN}(\cdot)$ indicates layer normalization, and $\text{Fun}(\cdot)$ represents either an attention network or a FFN. 

Following \cite{tang2020unbiased}, the frequency bias \cite{zellers2018neural} and the union features from the union boxes of two objects are added after HR-GAT, considering that calculating the frequency bias and union features is space-consuming. We use the binary cross-entropy loss for the relationship prediction.

\section{The Experiments}

\begin{table*}[t!]
\centering
\small
{
\begin{tabular}{lcccccccccc}
\toprule
	&\multicolumn{5}{c}{With Graph Constraint}&	\multicolumn{5}{c}{Without Graph Constraint}	\\	
Methods (\%)	&mR@50&	mR@100&	R@50&	R@100&Mean&	mR@50&	mR@100&	R@50&	R@100&Mean\\
		\hline
DR-GCN$^\Diamond$ \cite{zhang2020dual}	&	8.4	&	9.5	&	28.1	&	31.5	&	19.4	&	13.5	&	18.6	&	31.8	&	37.6	&	25.4	\\
GPS-Net$^\Diamond$ \cite{lin2020gps}	&	-	&	9.8	&	28.4	&	31.7	&	-	&	-	&	-	&	-	&	-	&	-	\\
IMP$^*$ \cite{li2017scene,tang2020unbiased}	&	4.6	&	5.9	&	27.2	&	32.5	&	17.6	&	5.7	&	8.6	&	28.7	&	35.9	&	19.7	\\
Motif$^*$ \cite{zellers2018neural,tang2020unbiased}	&	7.3	&	8.4	&	\textbf{32.9}	&	\textbf{37.1}	&	21.4	&	12.5	&	16.5	&	\textbf{36.9}	&	\textbf{43.5}	&	27.4	\\
VCTree$^*$ \cite{tang2019learning,tang2020unbiased}	&	7.0	&	8.0	&	31.8	&	35.9	&	20.7	&	12.2	&	16.5	&	35.8	&	42.3	&	26.7	\\
Motif-TDE$^*_d$ \cite{tang2020unbiased}	&	8.7	&	10.6	&	16.8	&	20.2	&	14.1	&	10.7	&	14.2	&	19.1	&	24.6	&	17.2	\\
VCTree-TDE$^*_d$ \cite{tang2020unbiased}	&	9.4	&	11.3	&	16.3	&	19.8	&	14.2	&	11.7	&	15.3	&	18.8	&	24.9	&	17.7	\\
Motif-cKD$_d$ \cite{wang2020tackling}	&	8.1	&	9.6	&	32.5	&	37.1	&	21.8	&	14.2	&	19.8	&	36.3	&	43.2	&	28.4	\\
VCTree-cKD$_d$ \cite{wang2020tackling}	&	7.7	&	9.1	&	32.0	&	36.1	&	21.2	&	13.9	&	19.0	&	35.9	&	42.4	&	27.8	\\
PCPL$_d^\Diamond$ \cite{yan2020pcpl}	&	9.5	&	11.7	&	14.6	&	18.6	&	13.6	&	10.4	&	14.4	&	15.2	&	20.6	&	15.2	\\
\hline
\textbf{HLN}	&	\textbf{11.3}	&	\textbf{13.1}	&	30.6	&	34.9	&	\textbf{22.5}	&	\textbf{17.6}	&	\textbf{23.1}	&	34.7	&	41.6	&	\textbf{29.3}	\\
\bottomrule
\end{tabular}
}
  \caption{Performance comparison on SGDet of the VG dataset. We also compute the mean over mR@50, mR@100, R@50, and R@100 to provide an overall evaluation. ``$^*$'' indicates the results of the method are calculated by this paper. ``-'' indicates the results are unavailable. ``$^\Diamond$'' means the corresponding method used Faster R-CNN with VGG-16. ``$_d$'' represents the unbiased SGG methods.}
\label{table1}
\end{table*}
\begin{figure*}[t!]
\centering
{\includegraphics[width=0.92\linewidth]{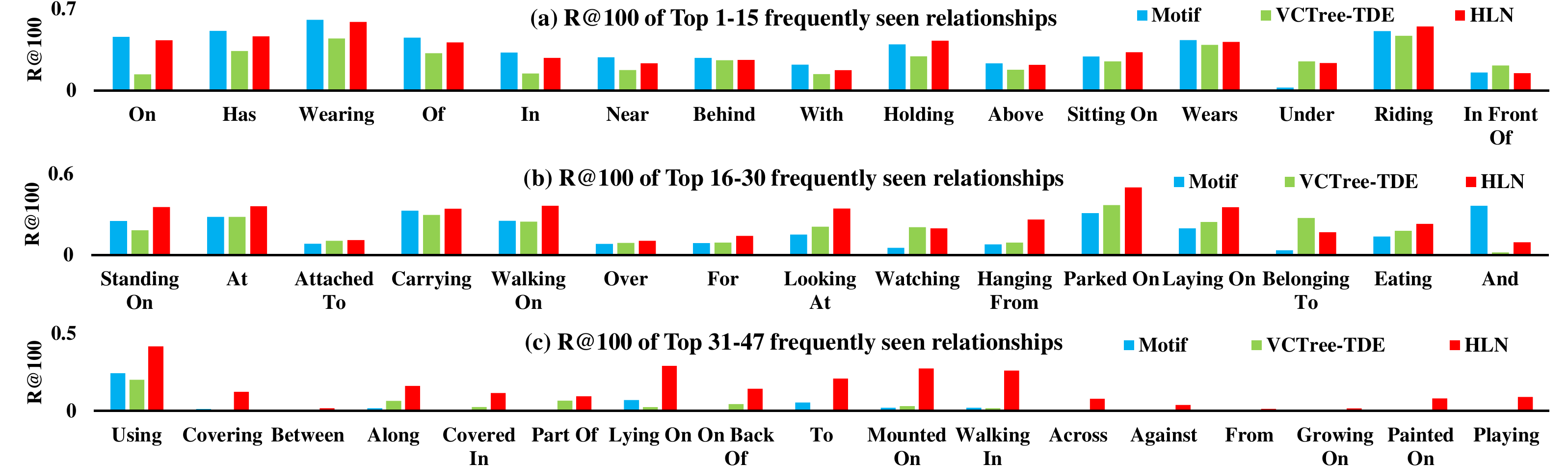}}
\caption{R@100 of Motif, VCTree-TDE, and HLN on SGDet without graph constraints of the top 1-47 frequently seen relationships of VG dataset. The least seen three relationships: ``Made Of'', ``Says'', and ``Flying In'', have too few labels to be detected by all methods. }
\label{fig3}
\end{figure*}

\begin{figure*}[t!]
\centering
{\includegraphics[width=0.92\linewidth]{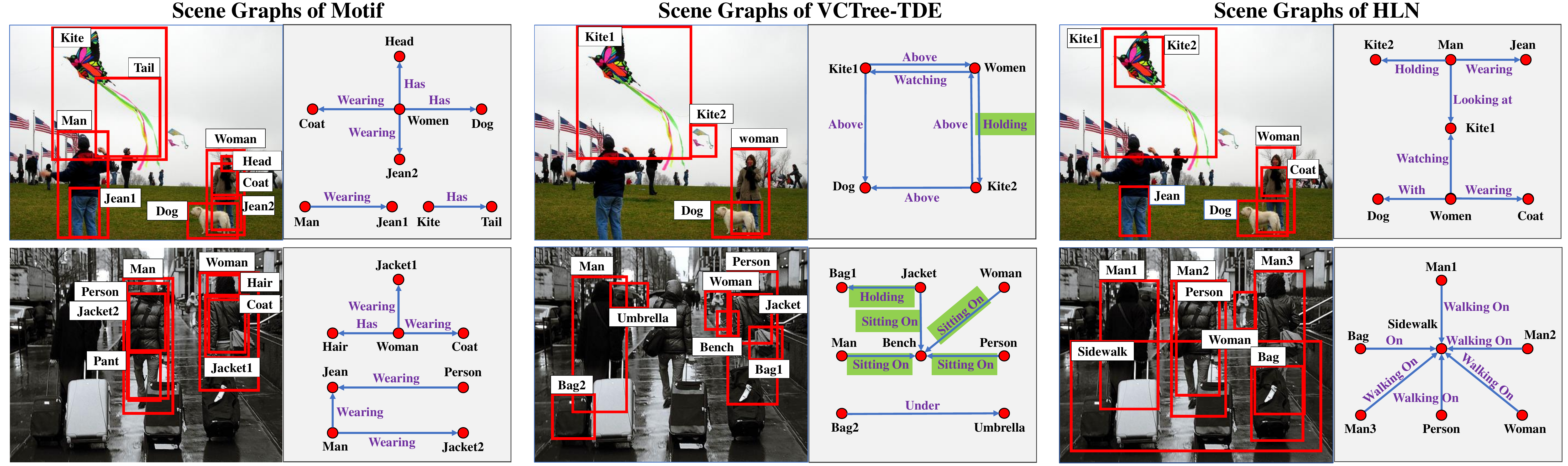}}
\caption{Scene graphs of two images that are generated by using the top-6 detected relationships of Motif, VCTree-TDE, and HLN.}
\label{fig4}
\end{figure*}

\subsection{Experimental Settings}\label{S4.1}
\textbf{Dataset.} We use the Visual Genome (VG) \cite{krishna2017visual} dataset to conduct all experiments. We notice multiple versions of VG datasets for SGG \cite{zhang2017visual,li2017scene,xu2017scene,zellers2018neural}. We select the most frequently used version \cite{zellers2018neural,tang2020unbiased}, which contains the most frequent 150 object categories and 50 predicate categories. We follow the experimental settings in \cite{zellers2018neural,tang2020unbiased}, which split the dataset into 70K/5K/32K as training/validation/test sets. 

\textbf{Evaluation Tasks.} We adopt two types of SGG tasks: scene graph detection (SGDet) and predicate classification (PreCls). SGDet generates scene graphs of images without extra-label information, whereas PreCls requires ground-truth objects. We mainly conducted experiments using the SGDet because SGDet is more practical than PreCls.

\textbf{Evaluation Metrics.}
We follow the precedent set of \cite{lu2016visual,xu2017scene} using recall as the evaluation metric. There are two types of recall: R@K and mR@K. R@K is calculated by averaging the recall of the top $K$ relationships of all images \cite{xu2017scene}. R@K demonstrates the correctness of generated relationships. mR@K is obtained by averaging the R@K independently calculated for each relationship (or predicate) category \cite{tang2019learning}. mR@K validates the ability of SGG methods to generate unbiased scene graphs. One object pair may contains multiple labeled predicates. Therefore, a increase of mR@K generally results in a drop of R@K. Considering the biased-distributed predicates in the test set \cite{chen2019knowledge,tang2019learning}, R@K and mR@K complementarily reflect the method's ability to generate correct and informative scene graphs.  

\textbf{Implementation Details.} We use the codebase and settings provided by~\cite{tang2020unbiased} to form a fair comparison with previous SGG and unbiased SGG methods. The backbone is Faster R-CNN based on ResNeXt-101-FPN \cite{lin2017feature}, and we use the same pre-trained model provided by~\cite{tang2020unbiased}. The layers before the ROIAlign are fixed during training. We optimize HLN by using the softmax cross-entropy loss for object classification and the binary cross-entropy loss for relationship prediction without debiasing strategies. The SGD optimizer with a momentum of 0.9 is adopted. The warm-up strategy \cite{vaswani2017attention} is also used to increase the learning rate from 0 to 0.001 in the first 5000 iterations. Then, the learning rate decays by 0.1 at 18,000 and 26,000 iterations with 34,000 iterations in total. The batch size is set to 12 during training. For each image, the top-64 object proposals are provided, and 256 relationship proposals, containing at most 25\% positive samples, are used to calculate the loss. 

\subsection{Performance Comparison} \label{S4.3}
\textbf{SGDet.} In Table \ref{table1}, we compare the proposed HLN with ten SGG methods on SGDet of the VG dataset. Both the results with and without graph constraints are provided for fair comparison. As shown in Table \ref{table1}, previous typical SGG methods perform well when using R@K but perform poorly when using mR@K, demonstrating their weak abilities to generate unbiased scene graphs. Unbiased SGG methods usually have high scores when using mR@K but have low scores when using R@K, because they severely reduce the detection of frequently-seen relationships. HLN performs the best when using mR@K while also maintaining relatively high scores when using R@K. HLN performs 20.2\%, 15.9\%, 87.7\%, and 76.3\% better than the best unbiased VCTree-TDE on mR@50, mR@100, R@50, and R@100 with graph constraints, respectively. HLN performs the best on the mean results. The above improvements show the great capabilities of HLN to generate unbiased scene graphs while guaranteeing the generated correctness.

An additional analysis to demonstrate the generalizability of HLN is shown in Fig.~\ref{fig3}, where Motif, VCTree-TDE, and HLN are compared using the top 1-47 seen relationship categories of the VG dataset (the top 1-15, the top 16-30, top 31-47 are shown in Fig.~\ref{fig3} (a), Fig.~\ref{fig3} (b), and Fig.~\ref{fig3} (c), respectively). Specifically, Motif and VCTree-TDE are compared because Motif achieves the best R@K and VCTree-TDE achieves the second-best mR@K in Table~\ref{table1}. We can draw the following conclusions: 1) Motif performs nearly the best on the top 10 frequently seen relationships as shown in Fig.~\ref{fig3}~(a). Since frequently seen relationships dominate the human-notated labels in the test set, Motif obtains the highest scores of R@K in Table~\ref{table1}. However, Motif performs poorly on detecting the remaining relationship categories and thus has low scores of mR@K; 2) VCTree-TDE outperforms Motif on most relationships shown in Fig.~\ref{fig3}~(b), demonstrating the usefulness of debiasing strategies. Nevertheless, current debiasing strategies severely reduce the detection of frequently seen relationships shown in Fig.~\ref{fig3}~(a) and cannot significantly improve the detection ability of informative relationships; 3) most relationships shown in Fig.~\ref{fig3}~(c) require integrating surrounding relationships, such as ``Playing''. We find that HLN performs the best on nearly all relationships shown in Fig.~\ref{fig3}~(b) and all relationships shown in Fig.~\ref{fig3}~(c), demonstrating that HLN has better relationship detection abilities than other methods. 

\begin{table}[t!]
\centering
\small
\resizebox{0.48\textwidth}{!}{
\begin{tabular}{lccccc}
\toprule
Methods (\%)	&	mR@50	&	mR@100	&	R@50	&	R@100&Mean	\\
\hline
IMP	\cite{li2017scene,tang2020unbiased}&	11.3	&	12.1	&	61.0	&	63.0	&	36.9	\\
Motifs \cite{zellers2018neural,tang2020unbiased}&	15.8	&	17.1	&	65.5	&	67.3	&	41.4	\\
VCTree \cite{tang2019learning,tang2020unbiased}&	16.8	&	18.0	&	\textbf{65.9}	&	\textbf{67.6}	&	42.1	\\
Motif-TDE$_d$ \cite{tang2020unbiased} &	24.5	&	28.0	&	44.5	&	49.8	&	36.7	\\
VCTree-TDE$_d$ \cite{tang2020unbiased}	&	\textbf{26.2}	&	\textbf{29.6}	&	42.4	&	46.6	&	36.2	\\
Motif-cKD$_d$ \cite{wang2020tackling}&	18.5	&	20.2	&	64.6	&	66.4	&	42.4	\\
VCTree-cKD$_d$ \cite{wang2020tackling}&	18.4	&	20.0	&	65.4	&	67.1	&	42.7	\\
\hline											
\textbf{HLN}	&	19.1	&	20.6	&	65.1	&	66.9	&	42.9	\\
\textbf{HLN+40Obj}	&	20.3	&	22.0	&	64.8	&	66.7	&	\textbf{43.5}	\\
\bottomrule
\end{tabular}
}
  \caption{Performance comparison on PreCls with graph constraints of the VG dataset. ``$_d$'' represents the unbiased SGG methods.}
\label{table2}
\end{table}

\textbf{Qualitative Comparison.} We further provide scene graphs of two images that are generated by using the top-6 detected relationships of Motif \cite{zellers2018neural,tang2020unbiased}, VCTree-TDE \cite{wang2019exploring,tang2020unbiased}, and HLN, respectively, in Fig.~\ref{fig4}. The green background indicates obviously incorrect relationships. In Fig.~\ref{fig4}, we can see that 1) Motif tends to generate frequently seen relationships, such as ``Wearing/Has'', instead of more informative and complex relationships, such as ``Looking At''. The above performance indicates that the generated scene graphs of most previous normal methods are far from practical. 2) VCTree-TDE can highlight less frequently seen relationships, such as ``Watching/Sitting On''. However, the generated correctness of VCTree-TDE is relatively low. There are many obvious mistakes, such as ``Jacket-Sitting On-Bench''. Most current debiasing strategies do not efficiently improve the detection ability. 3) HLN generates the most informative and correct scene graphs. Notably, HLN is capable of generating proper relationships to describe different objects in different situations. For example, in the street situation, HLN can distinguish ``Bag-On-Sidewalk'' and ``Person-Walking On-Sidewalk''. 

\textbf{PreCls.} Table~\ref{table2} compares the HLN with seven SGG methods on PreCls with graph constraints of the VG dataset. These methods are selected because they use the same codebase \cite{tang2020unbiased}. From Table~\ref{table2}, we can see that HLN performs the best regarding the mean performance. However, the improvements are not that significant compared with the results of SGDet. We believe this is because HLN integrates surrounding relationships, which requires numerous objects. For example, in Fig.~\ref{fig1} (b), HLN generates ``Lying On'' because it integrates multiple objects, including the boy's leg and head. However, these details are not always provided in the ground truth. In Table~\ref{table2}, we propose a variants: HLN+40Obj, which adds the top-40 detected objects with the human-notated objects for PreCls. HLN+40Obj outperforms HLN on mR@K. The improvements validate our above claim. Combining detected objects and human-notated objects to improve PreCls is a potential future work. 

\begin{table}[t!]
\centering
\small
\resizebox{0.48\textwidth}{!}{
\begin{tabular}{lccccc}
\toprule
Methods (\%)	&	mR@50	&	mR100	&	R@50	&	R@100	&	Mean	\\
\hline
SG-T \cite{yu2020cogtree}	&	7.81	&	9.26	&	\textbf{32.52}	&	\textbf{36.95}	&	21.64	\\
G-R \cite{yang2018graph}&	7.70	&	9.19	&	31.99	&	36.58	&	21.37	\\
RTN	\cite{koner2020relation}&	7.88	&	9.32	&	32.54	&	36.86	&	21.65	\\
HLN:O+R	&	7.66	&	9.03	&	32.50	&	36.87	&	21.52	\\
\hline
G-R+HR-GAT	&	8.54	&	10.33	&	31.44	&	35.98	&	21.57	\\
SG-T+HR-GAT	&	9.16	&	10.74	&	32.16	&	36.57	&	22.16	\\
\textbf{HLN}	&	\textbf{11.28}	&	\textbf{13.14}	&	30.56	&	34.89	&	\textbf{22.47}	\\
\bottomrule
\end{tabular}
}
\caption{Performance comparison of six methods on SGDet.}
\label{table5}
\end{table}

\begin{figure*}[t!]
\centering
{\includegraphics[width=0.93\linewidth]{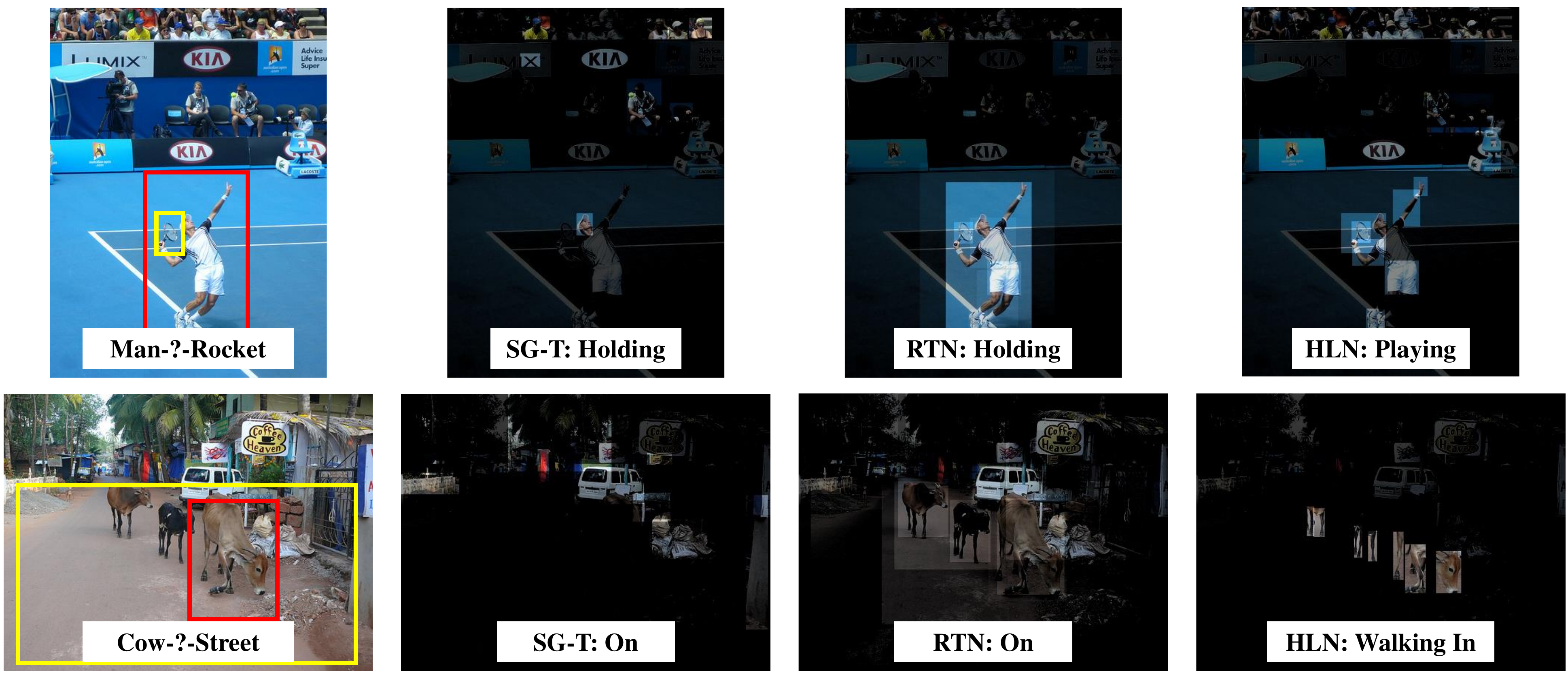}}
\caption{Attention Visualization of SG-T, RTN, and HLN of two informative relationships: ``Playing'' and ``Walking In''. }
\label{fig5}
\end{figure*}

\subsection{Discussion of OR-GAT and HR-GAT}
We discuss the benefits of OR-GAT and HR-GAT in this subsection. We reimplemented three recent state-of-the-art GATs for SGG: Graph R-CNN (G-R)~\cite{yang2018graph}, relation Transformer network (RTN)~\cite{koner2020relation}, and SG Transformer (SG-T)~\cite{yu2020cogtree}. Specifically, Graph R-CNN passes information between objects and their constituted relationships, RTN passes information between relationships, and SG-T passes information only from objects to relationships. Table~\ref{table5} provides the performance of SG-T, RTN, G-R, HLN:O+R, and HLN. Table \ref{table5} also provides SG-T+HR-GAT and G-R+HR-GAT, which adds one HR-GAT after the corresponding GATs. We observe that both SG-T, RTN, G-R, and HLN:O+R yield similar performance; when adding HR-GAT, the mR@K of all methods are significantly improved; and HLN achieves the state-of-the-art performance. The above results demonstrate that 1) current GATs still weak in modeling high-level connections between graph components such as transitive inference, 2) HR-GAT significantly improves the model reasoning ability via transitive inference, and 3) incorporating OR-GAT and HR-GAT leads to efficient and effective incorporation of the interactions and transitive inference between graph components. 

We provide attention visualization of SG-T, RTN, and HLN in Fig.~\ref{fig5} using two informative relationships: ``Playing'' and ``Walking In''.  Both SG-T and RTN provide less informative and straightforward relationships, \ie, ``Holding'' and ``On''. Besides, the attention of SG-T is relatively illogical, \eg, for ``Holding'', SG-T highlights the head of all existing persons. RTN's attention generally focuses on similar relationships, 
\eg, for ``On'', RTN highlights most ``Cow-On-Steet'' relationships. Only HLN can precisely detect both two complex relationships. The attention visualization demonstrates the reasoning and integrating processes of HLN: for ``Man-Playing-Rocket'', man's body parts and surrounding playgrounds are highlighted; for ``Cow-Walking In-Street'', the head and the legs of cow are highlighted.

\begin{table}[t!]
\centering
\small
\resizebox{0.48\textwidth}{!}{
\begin{tabular}{lccccc}
\toprule
Methods (\%)	&	mR@50	&	mR@100	&	R@50	&	R@100	&	Mean	\\
\hline
HLN:B	&	7.32	&	8.63	&	32.28	&	36.58	&	21.20	\\
HLN:O	&	7.68
	&   9.13
	&	32.32
	&   36.87
	&	21.50	\\
HLN:O+R	&	7.58
	&	9.14
	&	32.44
	&	36.92
	&	21.52	\\
\hline											
\textbf{HLN}	&	\textbf{11.28}	&	\textbf{13.14}	&	30.56	&	34.89	&	\textbf{22.47}	\\
\bottomrule
\end{tabular}}
  \caption{Performance comparison of HLN and three variants.}
\label{table3}
\end{table}

\begin{table}[t!]
\centering
\small
\resizebox{0.48\textwidth}{!}{
\begin{tabular}{lccccc}
\toprule
Methods (\%)		&	mR@50	&	mR@100	&	R@50	&	R@100	&	Mean	\\
		\hline
\multicolumn{6}{l}{Transformer Layer Number}											\\
0		&	10.77	&	12.76	&	30.57	&	35.05	&	22.29	\\
1		&	11.35	&	13.08	&	29.87	&	34.24	&	22.14	\\
3		&	10.97	&	12.98	&	30.64	&	35.09	&	22.42	\\
	\hline
\multicolumn{6}{l}{OR-GAT Number}												\\
0		&	8.47	&	9.95	&	\textbf{31.96}	&	\textbf{36.28}	&	21.67	\\
2		&	9.72	&	11.37	&	31.77	&	36.20	&	22.27	\\
	\hline
\multicolumn{6}{l}{Network Dimension}												\\
512		&	10.31	&	11.91	&	30.96	&	35.39	&	22.14	\\
1024		&	10.80	&	12.45	&	30.92	&	35.36	&	22.38	\\
	\hline
\multicolumn{6}{l}{Multi Head Number}							 					\\
4		&	11.19	&	 12.97	&	30.38 &34.82&22.34		\\
12		&	\textbf{11.72}	&	\textbf{13.64}	&	29.73	&	34.13	&	22.31	\\
	\hline
\multicolumn{6}{l}{With and Without Mask}	\\				
Without &  11.00	&	12.80	&	30.38	&	34.80	&	22.25	\\
\hline
\textbf{HLN} &	11.28	&	13.14	&	30.56	&	34.89	&	\textbf{22.47}	\\
\bottomrule
\end{tabular}
}
  \caption{Ablation experiments of HLN on SGDet.}
\label{table4}
\end{table}

\subsection{Ablation Studies}\label{S4.4}
\textbf{Interaction and Inference.} Table~\ref{table3} compares HLN and three variants to analyzes the beneficial effects of the interaction and transitive inference for HLN. In Table~\ref{table3}, all methods share the same OPN and object classifier. HLN:B is the baseline that does not use any interaction and inference. HLN:O and HLN:O+R only exploit object interaction based on Transformer layer and interaction between graph components of OR-GAT, respectively. Besides, HLN:O contains four layers of Transformer layers and HLN:O+R contains two OR-GATs to reduce the improvements from the larger computational operations when compared with HLN. We can observe that HLN:B performs the worst; HLN:O+R outperforms HLN:O; and HLN performs the best. These improvements indicate that all connections are beneficial to the SGG. HLN obtains the biggest improvements, which validates that transitive inference serves an significant consideration for relationship detection.

\textbf{Methodological Details.} Table \ref{table4} provides ablation experiments to analyze the methodological details of HLN from five components: the Transformer layer number, the OR-GAT number, the network dimension, the multi-head number, and whether exploiting the mask in OR-GAT.  The default settings of HLN are two Transformer Layer, one OR-GAT, 768 network dimension, eight multi heads, and using the mask of OR-GAT. All experiments were conducted by changing one component while fixing the others as the HLN. We can see that a proper setting of components benefits the detection of HLN.  
\vspace{-0.1cm}
\section{Conclusions}
\vspace{-0.1cm}

This paper proposes hyper-relationship learning network (HLN) to explore interactions of graph components and transitive inference of hyper-relationships for scene graph generation (SGG). Specifically, HLN models scene graphs as hypergraphs, and uses OR-GAT to combine object and relationship interaction, and HR-GAT to integrate transitive inference of hyper-relationships. Experimental results on the VG dataset demonstrate the superior reasoning and integrating abilities of HLN to detect informative relationships. HLN uses no unbiased strategies. Therefore, one of our future work is to combine HLN with unbiased SGG.




{\small
\bibliographystyle{ieee_fullname}
\bibliography{main}
}

\newpage
\newpage

\appendix
\maketitle

\section*{Overview of Appendixes}
This supplemental material provides more experiments and analysis of the hyper-relationship learning network (HLN). Specifically, we first explain the network complexity of HLN (Section \ref{S1}). Then, we demonstrate the capabilities of HLN by providing more comparison experiments (Section \ref{S2}). Next, we explain why we select both mean Recall and Recall as the evaluation metrics (Section \ref{S3}). Afterward, we discuss the difference between OR-GAT and other graph attention networks (GATs) designed for SGG (Section \ref{S4}). Last, we validate the reasoning and integrating abilities of HLN by visualizing and comparing the attention of HR-GAT and other GATs for different types of relationships (Section \ref{S5}).

\section{Network Complexity} \label{S1}
\begin{figure}[t]
{\includegraphics[width=1.0\linewidth]{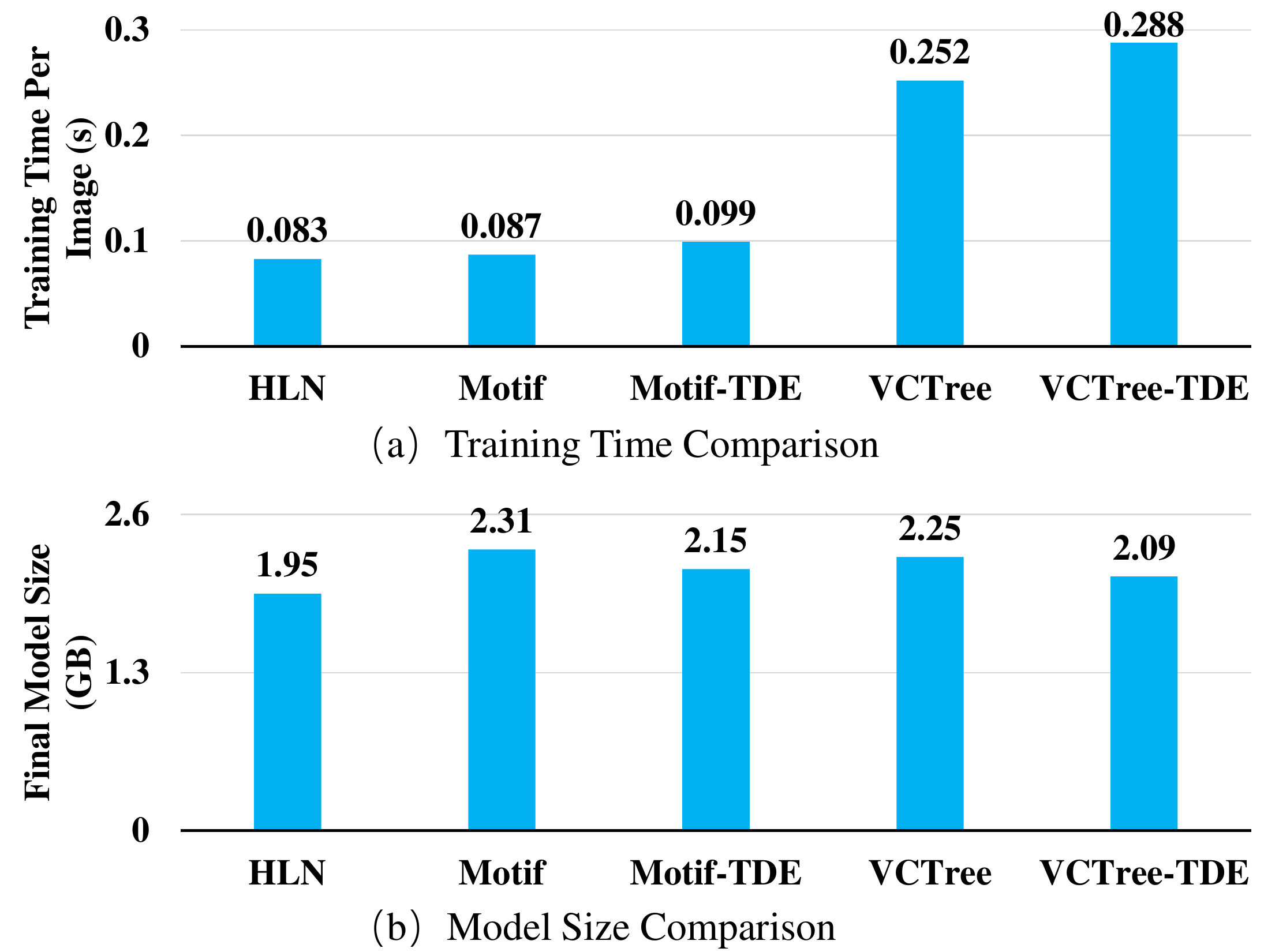}}
\caption{Complexity comparison of HLN, Motif \cite{zellers2018neural,tang2020unbiased}, Motif-TDE \cite{tang2020unbiased}, VCTree \cite{wang2019exploring,tang2020unbiased}, and VCTree-TDE \cite{tang2020unbiased}. (a) provides the training time per image. (b) presents the final model sizes. All methods were trained using the same codebase and nearly the same hyperparameters \cite{tang2020unbiased}.}
\label{fig6}
\end{figure}

This section presents the complexity of HLN. As mentioned in the main paper, HLN designs OR-GAT to reduce the space and time complexity of calculating relationship interaction based on attention networks. Specifically,  one OR-GAT first passes information from relationships to objects and then passes information from objects to relationships. In such a manner, during training, OR-GAT can implicitly exploit all objects and relationships to model their interactions. Furthermore, HLN only adds one HR-GAT at the end of the interaction processes considering the computational efficiency. During training, HR-GAT can use only 256 relationships as query and 256$\times$64$\times$4 corresponding relationships to constitute hyper relationships as key and value. Here, 256 relationships are used because we sampled 256 relationships to calculate the loss of relationship prediction. The sampled 256 relationships contain at most 64 positive samples. The time and space complexity during training are significantly reduced. Fig. \ref{fig6} compares the training time and model sizes of HLN with four state-of-the-art SGG and unbiased SGG methods: Motif \cite{zellers2018neural,tang2020unbiased}, Motif-TDE \cite{tang2020unbiased}, VCTree \cite{wang2019exploring,tang2020unbiased}, and VCTree-TDE \cite{tang2020unbiased}. All comparing methods are obtained based on the codebase \cite{tang2020unbiased} and use nearly the same experimental settings. Fig. \ref{fig6} shows that HLN has the least training time and smallest model sizes. The above performance validates that HLN has relatively lower training model complexity.

\section{Performance Comparison} \label{S2}

\begin{figure*}[t]
{\includegraphics[width=1.0\linewidth]{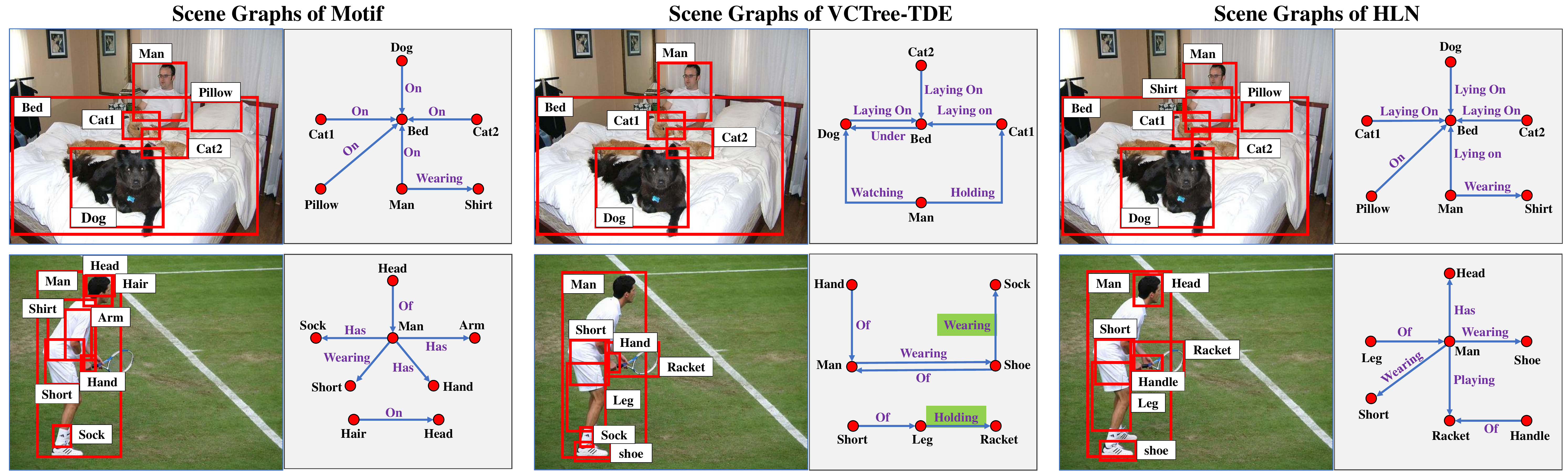}}
\caption{Scene graphs of two images that are generated by using the top-6 detected relationships of Motif \cite{zellers2018neural,tang2020unbiased}, VCTree-TDE \cite{wang2019exploring,tang2020unbiased}, and HLN.}
\label{fig7}
\end{figure*}

\begin{figure}[t]
{\includegraphics[width=1.0\linewidth]{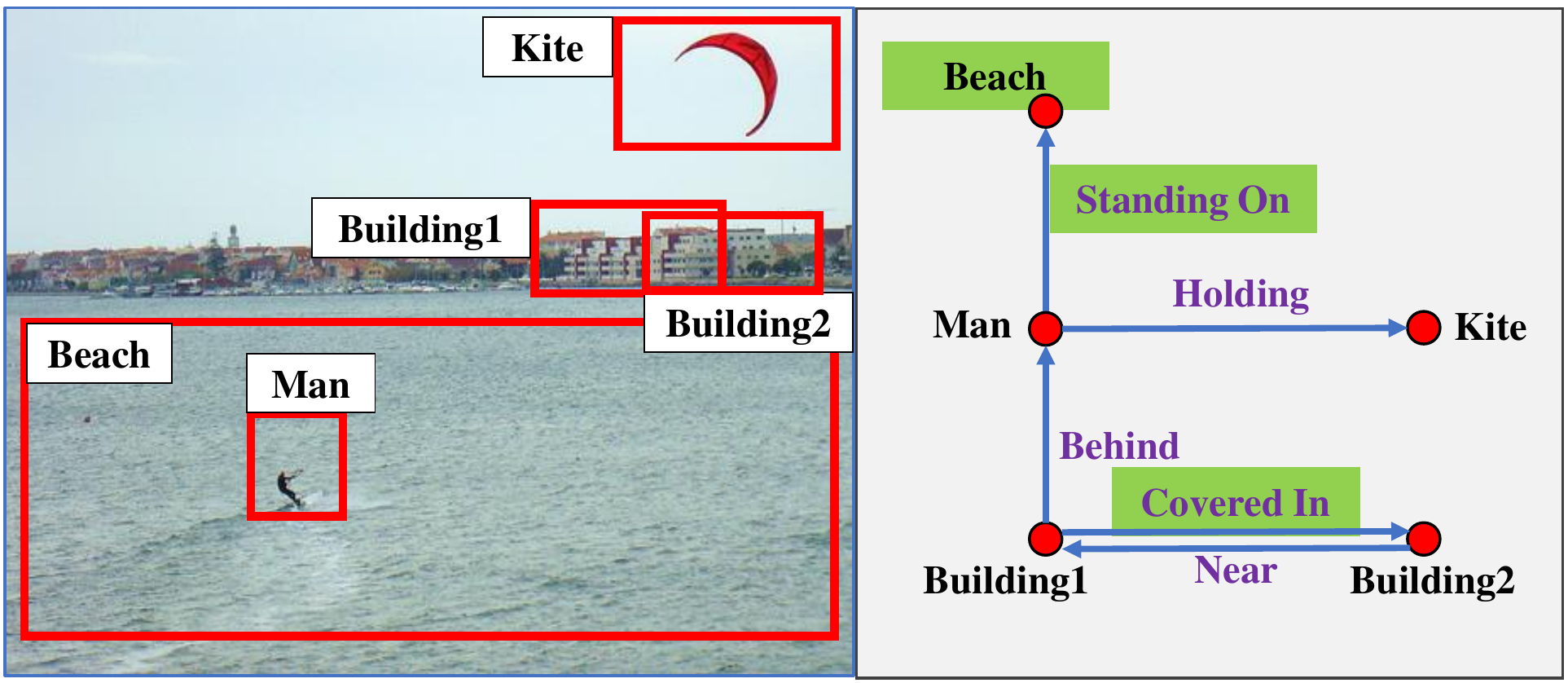}}
\caption{One negative scene graph that is generated by using the top-5 detected relationships of HLN.}
\label{fig8}
\end{figure}

\begin{table*}[t]
\centering
{
\begin{tabular}{lcccccccccc}
\toprule
	&\multicolumn{5}{c}{With Graph Constraint}&	\multicolumn{5}{c}{Without Graph Constraint}	\\	
Methods	&mR@50&	mR@100&	R@50&	R@100&Mean&	mR@50&	mR@100&	R@50&	R@100&Mean\\
		\hline
Motif \cite{zellers2018neural}	&	-	&	-	&	27.3	&	30.5	&	-	&	-	&	-	&	-	&	-	&	-	\\
VCTree \cite{tang2019learning}	&	6.9	&	8.0	&	27.9	&	31.3	&	18.5	&	-	&	-	&	-	&	-	&	-	\\
KERN \cite{chen2019knowledge}	&	6.4	&	7.3	&	27.1	&	29.8	&	17.7	&	11.7	&	16.0	&	\textbf{30.9}	&	\textbf{35.8}	&	23.6	\\
GPS-Net \cite{lin2020gps}	&	-	&	9.8	&	\textbf{28.4}	&	\textbf{31.7}	&	-	&	-	&	-	&	-	&	-	&	-	\\
PCPL$_d$ \cite{yan2020pcpl}	&	\textbf{9.5}	&	\textbf{11.7}	&	14.6	&	18.6	&	13.6	&	10.4	&	14.4	&	15.2	&	20.6	&	15.2	\\
Motif$^*$ \cite{zellers2018neural,tang2020unbiased}	&	6.1	&	7.2	&	25.5	&	28.6	&	16.9	&	10.3	&	14.2	&	28.6	&	33.7	&	21.7	\\
VCTree$^*$ \cite{tang2019learning,tang2020unbiased}	&	6.1	&	7.1	&	24.6	&	27.5	&	16.3	&	10.2	&	13.6	&	27.9	&	32.5	&	21.0	\\

Motif-TDE$_d^*$ \cite{tang2020unbiased}	&	4.9	&	5.9	&	24.6	&	27.9	&	15.8	&	7.4	&	10.7	&	26.9	&	31.8	&	19.2	\\
VCTree-TDE$_d^*$ \cite{tang2020unbiased}	&	4.5	&	5.4	&	23.1	&	26.2	&	14.8	&	6.5	&	9.1	&	25.3	&	29.9	&	17.7	\\
\hline
HLN	&9.2	&	10.8	&	26.2	&	30.1	&	\textbf{19.1}	&	\textbf{14.4}	&	\textbf{19.4}	&	29.6	&	35.2	&	\textbf{24.7}	\\

\bottomrule
\end{tabular}
}
  \caption{Performance comparison using SGDet on the VG dataset. ``$^*$'' indicates the method's results are calculated by this paper. The results of the rest methods are obtained based on their corresponding papers. ``-'' means the results are unavailable. $_d$ represents the method is designed for generating unbiased scene graphs. The best performance is highlighted in boldface.}
\label{table6}
\end{table*}

\textbf{Qualitative Comparison.}
Fig. \ref{fig7} provides scene graphs of two images that are generated by using the top-6 detected relationships of Motif \cite{zellers2018neural,tang2020unbiased}, VCTree-TDE \cite{wang2019exploring,tang2020unbiased}, and HLN, respectively. We use Motif and VCTree-TDE as the representative normal SGG and unbiased SGG methods, respectively. The Motif and VCTree-TDE are selected because they obtain the best R@K and the second-best mR@K when using Faster R-CNN with ResNeXt-101-FPN as the object proposal network (OPN). The green background indicates incorrect relationships. In Fig. \ref{fig7}, we can see that 1) Motif tends to generate frequently seen relationships, such as ``On/Wearing/Has/Of'' (the top-three frequently seen predicates in the training set), instead of more informative and complex relationships, such as ``Lying On/Playing''. The above performance indicates that the scene graphs generated by most previous SGG methods are still far from practical. 2) VCTree-TDE could highlight less frequently seen relationships, such as ``Watching/Laying On''. However, the generated correctness of VCTree-TDE is relatively low, partly because incorrect relationships, such as the ``Leg-Holding-Racket'' shown in Fig. \ref{fig7}, could be regarded as one type of less frequently seen or even unseen relationship. Current debiasing strategies of unbiased SGG solve the biased problems of training datasets but do not efficiently improve the detection ability of SGG methods. 3) HLN generates the most informative and correct scene graphs. For example, in the tennis court situation, HLN could detect relatively complicated relationships, such as the ``Man-Playing-Rocket‘’. Notably, HLN is capable of generating proper relationships to describe different objects concerning the objects' intrinsic characteristics. For example, in the bed situation, HLN could distinguish the difference between the dog and the pillow, thus generating ``Dog-Lying On-Bed'' and ``Pillow-On-Bed''.

Fig. \ref{fig8} provides one negative scene graph that is generated by using the top-5 detected relationships of HLN. We can observe that 1) if objects have been falsely detected, the corresponding relationships would be incorrect. Therefore, how to improve object detection also remains a significant consideration of HLN. 2) HLN generates the incorrect relationship: Building1-Covered In-Building2. However, intuitively, one building could not be covered in another building. In order to obtain more informative yet reasonable relationships, the common knowledge should be explored and exploited in HLN in the future.

\textbf{Quantitative Comparison Using VGG-16 Backbone.}
We additionally present the performance of HLN using Faster R-CNN with VGG-16 because many previous SGG methods adopt Faster R-CNN with VGG-16 as their object proposal networks (OPN). There are two types of commonly used pre-trained VGG-16 OPN: the Motif OPN \cite{zellers2018neural} and the RelDN OPN \cite{zhang2019graphical}. Unfortunately, the codebase of \cite{tang2020unbiased} has different architectures compared with the Motif OPN \cite{zellers2018neural}. The RelDN OPN was trained using both the training and validation datasets. The methods that adopted the RelDN OPN also used both training and validation datasets to train their relationship prediction \cite{zhang2019graphical,lin2020gps}. Therefore, we decided to train a Faster R-CNN with VGG-16 based on the codebase \cite{tang2020unbiased} using only the training dataset for a fair comparison. Specifically, most hyperparameters followed \cite{tang2020unbiased}. The Faster R-CNN with VGG-16 was initialized based on the model of ImageNet \cite{deng2009imagenet}. Image flip was used. Each batch contained 16 images. The learning rate was initialized as 0.01 and decayed by 0.1 at 37,500 and 52,500 iterations. The training lasted for 67,500 iterations. The final mAP@0.5 of object detection on the validation dataset is 21.2\%. Next, we fixed the parameters of OPN and only trained the HLN, mostly following the implementation details described in the main paper. The overlap requirements \cite{zellers2018neural} were also not considered, but the image flip was used. Each batch contained 12 images. The learning rate is initialized as 0.001 and decayed by 0.1 at 24,000 and 34,000 iterations. The training lasted for 40,000 iterations.

Table \ref{table6} provides the performance of HLN and seven comparing methods. Motif$^*$, VCTree$^*$, Motif-TDE$^*$, VCTree-TDE$^*$ are calculated by using the codebase \cite{tang2020unbiased} with our trained VGG-16 OPN. The results of the rest methods are obtained based on their corresponding papers. The best performance is highlighted in boldface. From Table \ref{table6}, we can draw the following conclusions: 1) Motif$^*$ and VCTree$^*$ have much lower performance compared with Motif and VCTree partly because different codebase details seriously influence the final detection performance. Besides, Motif$^*$ and VCTree$^*$ are trained only based on SGDet, whereas most of the previous papers firstly trained models based on PreDet and then fine-tuned their models based on SGDet \cite{zellers2018neural,wang2019exploring}. 2) Previous SGG methods achieved high scores using R@K but low scores using mR@K. These SGG methods ignore relationship connections, and they poorly recognize less-frequently seen relationships. 3) Unbiased SGG methods, except for the PCPL, perform unsatisfactorily no matter using R@K and mR@K. We believe this is because the low accuracy of object detection reduces most debiasing strategies' performance. The mAP@0.5 using ResNeXt-101-FPN of object detection of validation dataset is 26.4\%, which is much higher than the 21.2\% using VGG-16. PCPL performs well when using mR@K; however, its R@K is severely decreased. The above performance indicates that unbiased SGG methods probably overly detect less-frequently seen relationships regardless of their detected relationships' correctness. 4) HLN performs the best when using the mean, nearly the best when using mR@K, and satisfactorily when using R@K. The above improvements validate the capabilities of HLN for SGG. HLN can detect both frequently-seen and less frequently-seen relationships.

\section{Why use both Mean Recall and Recall?} \label{S3}
In the human-annotated relationships in the Visual Genome dataset, some object pairs are intuitively labeled as frequently-seen predicates instead of more informative predicates. For example, in Fig. \ref{fig7}, both ``Dog-On-Bed'' and ``Dog-Lying on-Bed'' are correct, but the ``On'' is more likely to be labeled as the predicate for the dog and the bed, even though the ``Dog-Lying on-Bed'' is more informative. Therefore, mean Recall generally provides a more fair evaluation than Recall. However, note that some predicates are more frequently seen is because these predicates commonly exist in images, such as ``Building-On-Ground'' and ``Pillow-On-Bed''. In these situations, using ``Sitting On'' or other so-called informative predicates may be redundant and incorrect. Therefore, the Recall also serves as a significant evaluation to test the correctness of the generated scene graphs of methods.

Nevertheless, only top-K relationships are selected as the final detected results of images. More detected informative relationships mean less detected frequently-seen relationships. It is natural for a method with high scores of mean Recall to have relatively low Recall scores (or with high scores of Recall to have relatively low mean Recall scores) concerning the bias labeling problem of ground truth. In summary, mean Recall and Recall are complementary to each other. A preferable SGG method should have both high mean Recall and Recall. We further compute the mean over mR@50, mR@100, R@50, and R@100 to provide an overall evaluation. 

\begin{table}[t]
\centering
{
\begin{tabular}{lccc}
\toprule
Methods	&	Type& Manner & Number	\\
\hline
G-R \cite{yang2018graph}&Obj. \& Rela.&Simu.& $O(N^3)$	\\
VSP \cite{zareian2020weakly}&Obj. \& Rela.&Simu.& $O(N^3)$	\\
SG-T \cite{yu2020cogtree}	&	Obj.	& None&$O(N^3)$\\
RTN	\cite{koner2020relation}&	Obj. \& Rela.& Alter.&$O(N^4)$	\\
\textbf{OR-GAT}	&	Obj. \& Rela.&Alter.&$O(N^3)$	\\
\bottomrule
\end{tabular}
  \caption{Summary of GATs exploiting interactions. ``Type'' indicates the interaction types. ``Obj.'' and ``Rela.'' represent the object and relationship, respectively. ``Manner'' represents the process of calculating interactions between object and relationships. ``Simu.'' and ``Alter.'' each indicate simultaneously and alternately updating the features of objects and relationships. 
  ``Number'' indicates the number of attention weight calculations in one GAT when giving $N$ objects. We suppose methods consider all objects and their relationships without using relationship selection processes.}
\label{table7}
}
\end{table}
\begin{table}[t!]
\centering
\resizebox{0.48\textwidth}{!}{
\begin{tabular}{lccccc}
\toprule
Methods (\%)	&	mR@50	&	mR100	&	R@50	&	R@100	&	Mean	\\
\hline
G-R \cite{yang2018graph}&	7.70	&	9.19	&	31.99	&	36.58	&	21.37	\\
VSP \cite{zareian2020weakly}& 7.91&9.31& \textbf{32.53}&36.94
&21.67\\ 
SG-T \cite{yu2020cogtree}&	7.81	&	9.26	&	32.52	&	\textbf{36.95}	&	21.64	\\
RTN	\cite{koner2020relation}&	7.88	&	9.32	&	32.54	&	36.86	&	21.65	\\
OR-GAT	&	7.66	&	9.03	&	32.50	&	36.87	&	21.52	\\
\hline
G-R+HR-GAT	&	8.54	&	10.33	&	31.44	&	35.98	&	21.57	\\
VSP+HR-GAT	&	8.50	&	10.16	&	31.99	&	36.43	&	21.77	\\
SG-T+HR-GAT	&	9.16	&	10.74	&	32.16	&	36.57	&	22.16	\\
\textbf{HLN}	&	\textbf{11.28}	&	\textbf{13.14}	&	30.56	&	34.89	&	\textbf{22.47}	\\
\bottomrule
\end{tabular}
}
\caption{Performance comparison of eight methods on SGDet.}
\label{table8}
\end{table}
\section{Why OR-GAT?} \label{S4}
We discuss the pros and cons of OR-GAT by comparing the network details of OR-GAT with four state-of-the-art GATs designed for scene graph generation; namely, G-R \cite{yang2018graph}, VSP \cite{zareian2020weakly}, RTN \cite{koner2020relation}, SG-T \cite{yu2020cogtree}. The details of all comparing GATs that use interactions of graph components are explained as follows.

\textbf{G-R.} The relationship features of G-R are initialized by integrating the information from the corresponding two objects. Then, the updated object features of G-R are obtained based on the corresponding relationships and other objects. Meanwhile, the updated relationship features are obtained based on the corresponding two objects.

\textbf{VSP.} The relationship features of VSP are randomly initialized by learnable embeddings. Then, the updated object features of VSP are obtained based on all relationships; meanwhile, the updated relationship features are obtained based on all objects.

\textbf{RTN.} RTN follows the encoder-decoder structure of the Transformer \cite{vaswani2017attention}. The relationship features of RTN are initialized by integrating the information from the corresponding two objects. Then, in the encoder, RTN exploits self-attention attention networks to update objects by modeling interactions between objects. Next, in the decoder, RTN first uses node-edge attention networks to pass information from objects to relationships and then uses an edge-edge self-attention network to pass information from updated relationships to updated relationships.  

\textbf{SG-T.} SG-T shares a similar attention process to RTN. Nevertheless, SG-T does not have the edge-edge self-attention network to pass information from relationships to relationships, for computational efficiency.  

\textbf{OR-GAT.} The relationship features of OR-GAT are initialized by integrating the information from the corresponding two objects. Then, the updated object features of OR-GAT are obtained based on the corresponding relationships. Next, the updated relationship features are obtained based on all updated objects. 

Table \ref{table7} provides a summary of all comparing GATs. We can observe that 1) RTN has the largest computational complexity because it directly calculates the relationship interaction based on relationship pairs. 2) SG-T does not exploit interactions between relationships because it only has networks to pass information from object to relationship. 3) G-R only updates relationship features based on the corresponding two objects. Such a kind of manner probably ignores the potential valuable interactions between the rest objects to the relationships \cite{ren2020scene}. 4) VSP updates object features based on all relationships. As we have shown in Table 5 of the main paper, without using a mask to limit the number of relationships, such a process obtains less performance.  5) OR-GAT updates object features based on the corresponding relationships and updates relationships based on the updated objects. Such a manner improves the performance experimentally. Besides, to implicitly obtain the interaction between relationships, OR-GAT only requires two GAT, which has less complexity than the four GATs of G-R and VSP, because G-R and VSP update relationship and object information based on the original relationship and object features.

Table \ref{table8} additionally provides the performance comparison of different methods. For robust performance, in Table \ref{table8}, G-R and VSP use all objects and relationships; the relationship features of VSP are initialized by integrating the information from the corresponding two objects. Table \ref{table8} also provides SG-T+HR-GAT, VSP+HR-GAT, G-R+HR-GAT, and OR-GAT+HR-GAT (namely, HLN), which add one HR-GAT after the corresponding GATs. We do not calculate RTN+HR-GAT because RTN+HR-GAT is too time- and space-consuming to be conducted using the same experimental settings as the rest methods. By adding HR-GAT after each comparing method, we could test the usefulness of each method's interaction information. From Table  \ref{table8}, we observe that both SG-T, VSP, RTN, G-R, and OR-GAT yield similar performance, but OR-GAT has relatively lower scores; when adding HR-GAT, the mR@K of all methods are significantly improved; and HLN achieves the state-of-the-art performance. Besides, both G-R and VSP update object and relationship features simultaneously. OR-GAT first updates object features and then updates relationships. The higher performance of HLN indicates that OR-GAT has a better ability to incorporate interactions between graph components. 

\section{Attention Visualization} \label{S5}
\begin{figure*}[t]
{\includegraphics[width=1.0\linewidth]{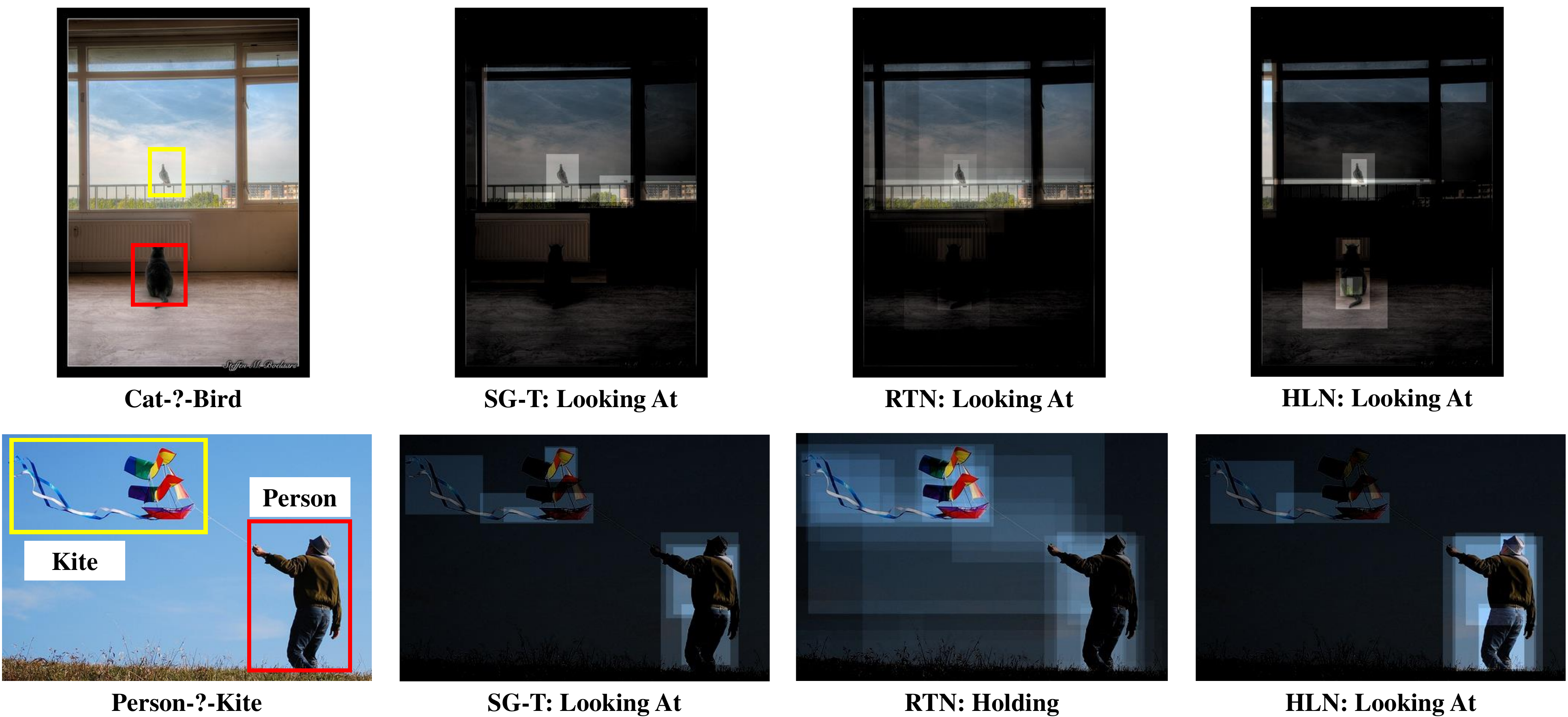}}
{\includegraphics[width=1.0\linewidth]{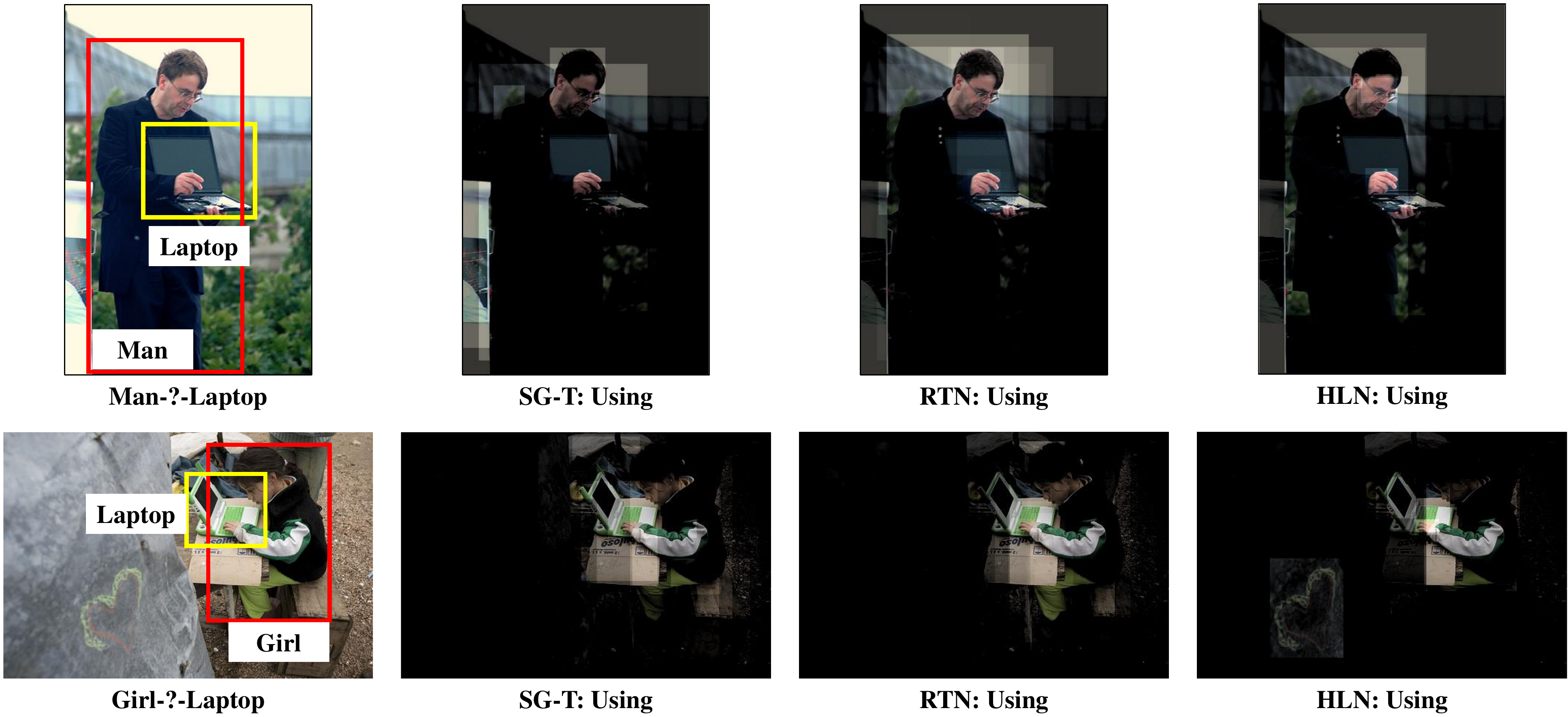}}
\caption{Attention Visualization of SG-T, RTN, and HLN using two types of relationships: ``Looking At'' and ``Using''. For each type of relationship, we provide two examples. From left to right, each column indicates the original image, attention of SG-T, attention of RTN, and attention of HLN. The corresponding detections are provided under each subfigure.}
\label{fig9}
\end{figure*}

\begin{figure*}[t!]
\includegraphics[width=1.0\linewidth]{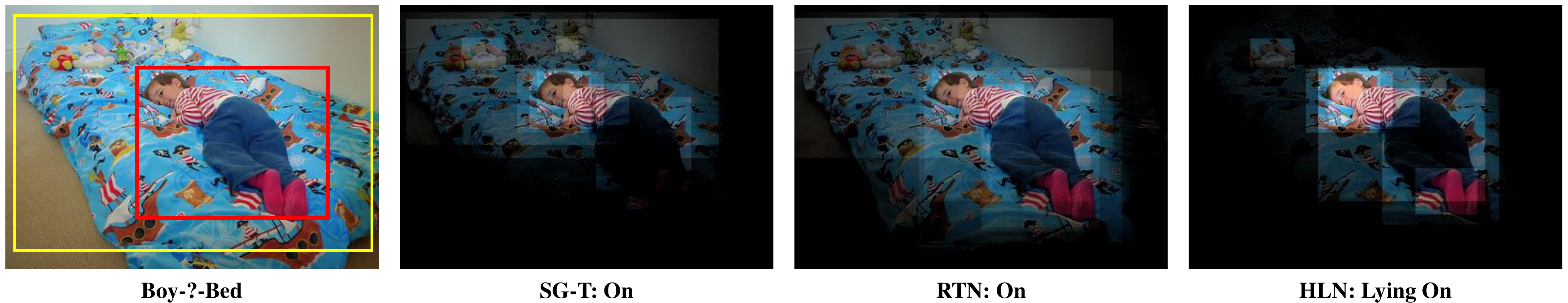}
\caption{Attention Visualization of SG-T, RTN, and HLN using ``Lying On''. From left to right, each column indicates the original image, attention of SG-T, attention of RTN, and attention of HLN. The corresponding detections are provided under each subfigure.}
\label{fig10}
\end{figure*}
We demonstrate the reasoning and integrating abilities of HLN by analyzing the attention of the HR-GAT in HLN when dealing with different relationships. For better visualization, for each relationship, we only highlight the top-10 objects of HR-GAT and the node-edge attention network of SG-T and the top-100 relationships (\ie, 10$\times$10 object pairs) of the edge-edge attention network of RTN. 

Fig. \ref{fig9} provides attention visualization of SG-T, RTN, and HLN using two types of relationships: ``Looking At'' and ``Using''.  We can see that although SG-T, RTN, and HLN generate correct relationships in most situations, HLN's attention obtains the best reasoning and integrating processes. For example, when generating ``Looking At'', only HLN highlights the heads of the cat and the man. When generating ``Using'', only HLN focuses on the hands of the man and the girl.  Fig. \ref{fig10} additionally provides attention visualization of SG-T, RTN, and HLN using one relatively complex relationship: ``Lying On''.  Both SG-T and RTN provide less informative and straightforward relationships, \ie, ``On''. Only HLN detects the correct relationship ``Lying On''. From Fig. \ref{fig10}, we can see that both SG-T and RTN focus on the bed and the boy. Only HLN concentrates on the essential details: the boy’s head and the boy’s leg.The above comparison validates HLN has a better reasoning ability to understand the connections of different components when detecting relationships.

\end{document}